# Anomaly Detection of UAV State Data Based on Single-class Triangular Global Alignment Kernel Extreme Learning Machine

**Feisha Hu[1], Qi Wang[1*], Haijian Shao[1, 2], Shang Gao[1] and Hualong Yu[1]**

[1] School of Computer, Jiangsu University of Science and Technology, Zhenjiang 212100, China
[2] Department of Electrical and Computer Engineering, University of Nevada, Las Vegas, NV 89154, USA
[*]Corresponding Author: Qi Wang. Email: wangqi@just.edu.cn

**Abstract:** Unmanned Aerial Vehicles (UAVs) are widely used and meet many demands in military and civilian fields. With the continuous enrichment and extensive expansion of application scenarios, the safety of UAVs is constantly being challenged. To address this challenge, we propose algorithms to detect anomalous data collected from drones to improve drone safety. We deployed a one-class kernel extreme learning machine (OCKELM) to detect anomalies in drone data. By default, OCKELM uses the radial basis (RBF) kernel function as the kernel function of the model. To improve the performance of OCKELM, we choose a Triangular Global Alignment Kernel (TGAK) instead of an RBF Kernel and introduce the Fast Independent Component Analysis (FastICA) algorithm to reconstruct UAV data. Based on the above improvements, we create a novel anomaly detection strategy FastICA-TGAK-OCELM. The method is finally validated on the UCI dataset and detected on the Aeronautical Laboratory Failures and Anomalies (ALFA) dataset. The experimental results show that compared with other methods, the accuracy of this method is improved by more than 30%, and point anomalies are effectively detected.

## 1 Introduction

Detect and predict the data generated by equipment, especially the detection of abnormal data. In recent years, in the industrial field, the production, manufacturing, and daily production of new energy represented by wind energy[1-3] and intelligent equipment represented by unmanned systems management have gradually highlighted its essential value and research significance.

The Unmanned Aerial Vehicles (UAVs) is a reusable unmanned aircraft normally controlled by the ground control center (GSC) or steered by an onboard program to achieve flight [4]. The application range of UAVs is extensive. For example, in military applications, UAVs can perform various tasks; in civilian applications, UAVs can meet the needs of aerial photography, express transportation, etc. In this context, UAVs have been studied more and more deeply. Such as, Lin et al. [5] proposed a Global Energy Efficiency Maximization (GREEN) strategy for supporting multi-UAV communication systems; Zhao et al. [6] proposed an SDN-enabled UAV-assisted vehicle computing offload Optimize the framework to minimize the system cost of vehicle computing tasks; Lin et al. [7] proposed an adaptive UAV deployment scheme to solve the coverage problem of UAV-assisted ground node (GNs) communication.

However, UAVs are limited by their conditions, such as size, weight, and cost, and compared with handled devices, UAVs still need to be improved in real-time perception and rapid decision-making, which leads to the safety and reliability of UAVs. There is a big gap between the sex and the crewed aircraft. Moreover, the application of UAVs is pervasive, which leads to the demands and uses generated in multiple types of complex environments, which also pose particular challenges to the safety of UAVs, these challenges mainly include the loss of UAV devices such as sensors, batteries, and motors in complex environments, and when the loss is too large, these devices will fail. In order to cope with this series of challenges, strengthening the detection of abnormal data of the drone itself and early warning and handling of potential problems will be more conducive to prolonging the service life of the drone and improving its adaptability to complex environments.

As the flight control part of the core control system of the UAV, the flight control system integrates a large number of multi-type sensors, and the data generated can more intuitively reflect whether the real-time state of the UAV is in good condition. The flight control system returns sensor data to the ground control center, and GSC can make an in-depth analysis and evaluation of the UAV's flight status based on these data. Many sensor-based anomaly detection methods have emerged in recent years, mainly divided into model-based methods [8] and data-driven methods [9].

Model-based methods use the physical model of the target system to detect anomalies. Such methods usually achieve better performance in terms of detection accuracy. However, establishing an accurate physical mode for each UAV system is often tricky. The applicability and anti-interference ability of the method is highly dependent on the constraints, such as the scene; the data-based method mass of data generated by the UAV sensor and uses the sensor data to detect the abnormal situation of the UAV.

Data-driven methods are mainly divided into two categories: supervised detection modus and unsupervised detection modus. Supervised detection methods deploy datasets mixed with normal and anomalous data to train models and achieve accurate anomaly detection. Although the anomaly detection method based on supervised learning can achieve accurate detection performance, it also has some problems. First, the anomaly detection method based on supervised learning needs to mix the training set of regular and abnormal data, and the corresponding labels when training the model, so engineers need to spend much energy to label the data set; secondly, because the trained model only learns The abnormal data type contained in the training set, so the model may not be able to identify the new abnormal data type, so the model needs to retrain the training set containing the abnormal data type. Unsupervised detection methods establish a boundary that includes all possible standard data points and use it as a basis for determining the nature of the test points. Since the data of UAVs usually have no label information, and the amount of abnormal data is limited and difficult to obtain, it is reasonable to choose an unsupervised detection essentials to detect anomalies in UAVs.

The innovations of this paper are as follows:

(1) First, considering the sparseness of abnormal data of UAVs, a single-class kernel extreme learning machine [10] (OCKELM) is selected as the basic model of this paper, the model can only use normal data to train the model, and can effectively detect the abnormal situation of the UAV.

(2) In addition, to improve the accuracy of OCKELM, the triangular global alignment kernel function [11] (TGAK) is used to replace the radial basis kernel function (RBF), and it is introduced into the OCKELM model to obtain a new model TGAK-OCELM. It has been verified that TGAK is a positive semi-definite kernel, so it satisfies Mercer's theorem. Comparing experiments with several typical unsupervised detection methods that have been approached via deploying the public datasets, the experiments show that the accuracy of TGAK-OCELM is higher than several other unsupervised detection methods.

(3) Finally, the UAV data features are extracted by manual extraction, and the features are reconstructed using Fast Independent Component Analysis [12] (FastICA). After testing, the reconstructed data improves the performance of the model.

The rest of this article is organized as follows. In Chapter 2, some UAV anomaly detection methods are introduced. Chapter 3 proposes a single-class extreme learning machine based on a triangular global alignment kernel, and independent component analysis is introduced to reconstruct the data. Chapter 4 illustrates the experimental environment and model evaluation metrics and then presents the experimental results to verify the effectiveness of the proposed method. The fifth chapter summarizes the full text.

## 2 Related Work

This section reviews the methods used in anomaly detection for drones. According to the types of algorithms designed in the research articles, there are three kinds of categories: model-based, supervised learning-based, and unsupervised learning-based methods.

### *2.1 Model-based UAV anomaly detection*

In the model-based approach, Rago et al. [13] utilized the interactive multi-model (IMM) Kalman filter method for fault detection and identification of Eagle Eye UAV sensors/actuators. For the F-16 fighter jet, Hajiyev et al. [14] proposed an Extended Kalman Filter (EKF) to detect the failure of sensors on the aircraft. Cork et al. [15] used nonlinear aircraft dynamics models, combining interactive multi-models and unscented Kalman filters to improve state estimation when inertial sensors fail. Gao et al. [16] proposed a fault detection method for tiltrotor UAV actuators based on an extended Kalman filter and multi-model adaptive estimation (MMAE), using EKF to estimate the state of the actuator data. Then MMAE is used to assign a conditional probability to each actuator and then to judge the failure condition. Bu et al. [17] combined the particle filter with the fuzzy inference system, utilized the particle filter to estimate the state of the sensor, and dispatched the difference between the state-estimated value and the GPS measurement value as input to the inference system, and then reasoned system to determine the degree of abnormality.

This type of method estimates the residual change of the system state by constructing a model of a specific system, so as to detect abnormal conditions in the system and often achieve good performance. Since the model of the target system needs to be used, when detecting abnormal conditions of other systems, the situation will get worse.

### *2.2 UAV anomaly detection based on supervised learning*

In supervised learning-based methods, Ge et al. [18] used the least squares support vector machine to predict UAV anomaly data points, while Bronz et al. [19] and Baskaya et al. [20] classified data from drones. While Arthur et al. [21] used self-learning (STL) multi-class support vector machines to detect signal spoofing anomalies and jam attacks against light UAVs. Titouna [22] divided the sensors into external sensors and internal sensors according to the sensor type and then used the KL divergence to calculate the difference between the data points of the external sensors to detect the abnormal situation of the UAV and used ANN to analyze the data of the internal sensors. Points are classified to distinguish anomalous data from drones. Nanduri [23] uses a recurrent neural network (RNN) with a long short-term memory (LSTM) and gated recurrent unit (GRU) structure to detect abnormal data, which does not require dimensionality reduction on the data, is more sensitive to short-term anomalies, and Potential failures can be detected.

Although this type of method can achieve good results in detecting abnormal data, it needs to acquire prior knowledge of abnormal data, that is, it needs to acquire each type of abnormal data to train the model. Therefore, it cannot identify unknown anomaly types, resulting in lower detection performance..

### *2.3 Unsupervised learning-based UAV anomaly detection*

Lin et al. [24] proposed a Mahalanobis distance-based UAV anomaly detection method among the methods based on unsupervised learning methods. The authors of the method determined the difference between different sensors by identifying the existence of non-statistically independent variable subgroups. And then use Mahalanobis distance to identify outliers in variable groups. Khalastchi et al. [25] split the data into datasets of several related attributes and then used Mahalanobis distance to detect anomalies. However, the numerical values calculated by such distance-based methods are unstable and depend on the setting of the threshold. Yong et al. [26] proposed an anomaly detection method based on Kernel Principal Component Analysis (KPCA) algorithm because of the dynamic UAV sensors and the high dimension of sensor data. KPCA was used to extract principal component vectors from sensor data and pass probability distribution to determine whether the drone was abnormal. Khan et al. [27] used the isolated forest algorithm to segment data points according to the eigenvalues of sensor data and generated a tree. The closer the data points to the root node of the tree, the higher the probability of abnormal points. Isolation forests do not use distance and density metrics to detect anomalies, which saves a lot of time, but requires a mixture of normal and anomalous data at training time. Whelan et al. [28] used various single-class classifiers to research UAV data, among which they employed a single-class support vector machine (OCSVM), local outlier factor algorithm (LOF), and autoencoder. The single-class SVM forms the boundary by average training data, and then the test data is standard if it is within the boundary, otherwise abnormal data. LOF judges data point anomalies by comparing the density between data points. Although the LOF is not affected by the data distribution, it is sensitive to the density parameter. The autoencoder generates a model by training standard data and then sends the test data as input to the model to get the output and compares the difference between the input and output to judge the abnormality of the test data. Chriki et al. [29] used a pre-trained convolutional neural network (CNN) and two manual methods, histogram of oriented gradients (HOG) and HOG3D, to extract useful features from videos collected by drones, and then applied OCSVM to detect anomalies data. Park [30] proposed a fault detection model based on stacked autoencoders for fault detection on UAV data. Guo [31] et al. combined the local density method with OCSVM to ameliorate the abnormal data detection performance of OCSVM by adjusting the tolerance of the decision boundary for outliers according to the degree of anomaly represented by the local density. Fu et al. [32] operated to filter and clustering algorithms and local density algorithms to perceive outliers. Avola et al. [33] used OCSVM as an anomaly detector for video surveillance at low altitudes of UAVs to detect the texture features of items in the video, and detect unknown objects according to different texture features. In recent years, deep learning technology has developed rapidly, and a large number of algorithms with excellent performance have appeared [34-36] and are practiced in all walks of life, including deep learning technology applied in the field of unmanned aerial vehicles. Jin et al. [37] proposed an anomaly detection model based on Transfomer for anomaly detection of videos shot in the air by UAVs, treating consecutive video frames as a sequence, using the encoder to learn features from the sequence, and then using the decoder. to predict the next frame. Tlili et al. [38] utilized two LSTM autoencoder models to train data on normal flight of unfaulted and unattacked UAVs, respectively, and then integrated the two models by setting a threshold to obtain a new model, which It can detect the flight status of the drone when it fails and the flight status of the drone when it is attacked. As a single-class classifier, OCSVM and autoencoder only need normal data for training, and can achieve good performance, but iteratively update parameters during the training process.

As a single-class classifier, OCKELM has the same function as OCSVM and autoencoder, and uses normal data to train the model. But the advantage of OCKELM is that for the weight, it does not need to be iteratively calculated but is determined at one time by solving the equation system.

## 3 Proposed Scheme

In this section, first, we briefly introduce OCKELM and TGAK, then propose OCKELM based on TGAK, namely TGAK-OCELM, and finally, introduce the FastICA method for reconstructing UAV data and introduce the UAV anomaly data detection process.

### *3.1 TGAK-OCELM*

Given $N$ training sets $\boldsymbol{T} = \{(x_i, y_i) \mid x_i \in R^d, y_i \in R, i = 1,2,3,\dots,N\}$, where $x_i$ is a sample instance with $D$ dimension, $y_i$ is the corresponding target tag and is a constant. In this paper, $y_i$ is set to 1.

The objective function of OCKELM is:

$$f(x) = h(x)^T \boldsymbol{\beta} \tag{1}$$

The optimization problem of OCKELM is defined as follows:

$$\begin{cases} min\ \dfrac{1}{2}\left\|\boldsymbol{\beta}\right\|_2^2 + \dfrac{C}{2}\sum_{i=1}^{N}\left\|e_i\right\|_2^2 \\ s.t.\ h(x_i)^T\boldsymbol{\beta} = y_i - e_i \end{cases} \tag{2}$$

where $\boldsymbol{\beta} = [\beta_1, \beta_2, \dots, \beta_l]^T$ is the output weight vector of the model, $l$ is the number of hidden cells of the model, $and\ e_i$ represents the instance training error of $x_i$. At the same time, the example mapping function of $x_i$ is represented by $h(x_i) = [h_1(x_i), h_2(x_i), \dots, h_l(x_i)]^T$, $\|\cdot\|_2^2$ represents the two norms of the vector, and $C$ is the regularization parameter, which is used to adjust the ratio of the output weight norm to the training error.

According to the KKT [39] theory, the Lagrange function of the optimization problem is:

$$L = \frac{1}{2}||\boldsymbol{\beta}||_2^2 + \frac{C}{2}\sum_{i=1}^{N}||e_i||_2^2 - \sum_{i=1}^{N}\alpha_i\left(h(x_i)^T\beta - y_i + e_i\right) \tag{3}$$

In Formula (2), once $\beta$, $e_i$, $\alpha_i$ discover the partial derivative and construct the result amount to 0, then we can obtain:

$$\begin{cases} \dfrac{\partial L}{\partial \beta} = \boldsymbol{\beta} - \sum_{i=1}^{N}\alpha_i\, h(x_i)^T = 0 \\ \dfrac{\partial L}{\partial e_i} = Ce_i - \alpha_i = 0 \\ \dfrac{\partial L}{\partial \alpha_i} = h(x_i)^T\boldsymbol{\beta} - y_i + e_i = 0 \end{cases} \tag{4}$$

Further derivation of the Formula (3) shows that:

$$\begin{cases} \boldsymbol{\beta} = \sum_{i=1}^{N}\alpha_i\, h(x_i)^T \\ e_i = \dfrac{1}{C}\alpha_i \\ h(x_i)^T\boldsymbol{\beta} - y_i + e_i = 0 \end{cases} \tag{5}$$

According to Formula (4), it can be concluded that:

$$\boldsymbol{\beta} = H^T\left(\frac{1}{C}I + HH^T\right)^{-1}\boldsymbol{Y} \tag{6}$$

where, $\boldsymbol{Y} = [y_1, y_2, \dots, y_N]$, $H = [h(x_1), h(x_2), \dots, h(x_N)]$. According to Mercer's condition, the kernel matrix $\Omega = HH^T$ is defined, where $\Omega(i,j) = K(i,j) = h(x_i)h^T(x_j)$, $i,j = 1, 2, \dots, N$, and $K(i,j)$ is a kernel function. The kernel function is generally selected as the radial basis function (RBF) kernel function $K(i,j) = exp(-\frac{||x_i - x_j||_2^2}{2\sigma^2})$, of RBF kernel function $\sigma$, is the width parameter of the function. From the Equation (1) and (6), the output $f(x)$ of the ockelm model can be deduced as:

$$\begin{aligned} f(x) &= h(x)^T\boldsymbol{\beta} \\ &= h(x)^T H^T\left(\frac{1}{C}I + HH^T\right)^{-1}\boldsymbol{Y} \\ &= \begin{bmatrix} K(x, x_1) \\ \vdots \\ K(x, x_N) \end{bmatrix}^T \left(\frac{1}{C}I + \Omega\right)^{-1}\boldsymbol{Y} \end{aligned} \tag{7}$$

When the training set $T$ is taken as the input, the output $\boldsymbol{Otr}$ of the model is:

$$\boldsymbol{Otr} = \Omega\left(\frac{1}{C}I + \Omega\right)^{-1}\boldsymbol{Y} \tag{8}$$

The distance between the output $\boldsymbol{Otr}$ of the training set and the target label $\boldsymbol{Y}$ is: $\boldsymbol{D} = |\boldsymbol{Otr} - \boldsymbol{Y}|$. $\boldsymbol{D}$ represents the difference between the output of the training set and the target label, the larger $\boldsymbol{D_i}$, the greater the error of example $i$. Sort the values in $\boldsymbol{D}$ in descending order to obtain $\widehat{\boldsymbol{D}}$, and set the instance as the threshold of abnormal value $\delta$ for:

$$\delta = \widehat{\boldsymbol{D}}(\lfloor \theta * N \rfloor) \tag{9}$$

$\delta$ represents the $\lfloor \theta * N \rfloor$ th error in the error $\widehat{\boldsymbol{D}}$ as the detection threshold. $\theta$ represents the anomaly proportion of selected training instances as anomalous instances. Therefore, the model output $\boldsymbol{Ot}$ for the *t*-th test instance $x_t$ is:

$$Ot = \begin{bmatrix} K(x_t, x_1) \\ \vdots \\ K(x_t, x_N) \end{bmatrix}^T \left(\frac{1}{C} I + \Omega\right)^{-1} \boldsymbol{Y} \tag{10}$$

Then, the distance between the output result $\boldsymbol{Ot}$ of the test instance and the target label $y$ is $Dt = |Ot - y|$, and the category corresponding to the final test instance $x_t$ can be defined as:

$$sign(\delta - Dt) = \begin{cases} 1, & x_t \textit{ is the normal data} \\ -1, & x_t \textit{ is the abnormal data} \end{cases} \tag{11}$$

Equation (7) indicates that if the value of $\delta - Dt$ is greater than 0, the label of $x_t$ is 1, which means that the test instance $x_t$ belongs to standard data. Otherwise, the test instance $x_t$ belongs to abnormal data. Algorithm 1 gives the pseudocode of the OCKELM algorithm.

**Algorithm 1 OCKELM**

Enter:

Training set: **T**, Test instance: $x_t$, Regularization parameter: *C*, Radial basis function parameter: $\sigma$, Outlier ratio: $\theta$

Output:

Label corresponding to $x_t$

(1) The kernel matrix$\Omega$is calculated by the radial basis (RBF) kernel function $K(i,j) = exp(-\frac{||x_i - x_j||_2^2}{2\sigma^2})$.

(2) The model output $\boldsymbol{Otr}$ of the training set is calculated by Formula (8).

(3) The abnormality threshold $\delta$ is calculated by Formula (9).

(4) Calculate the kernel function $\begin{bmatrix} K(x_t, x_1) \\ \vdots \\ K(x_t, x_N) \end{bmatrix}^T$ of the test instance $x_t$ and the training set **T** via employing the radial basis (RBF) kernel function $K(i,j) = exp\left(-\frac{\left\|x_i - x_j\right\|_2^2}{2\sigma^2}\right)$.

(5) The model output $\boldsymbol{Ot}$ of the instance $x_t$ is calculated by Equation (10).

(6) The category of the instance $x_t$ is determined by Equation (11).

Cuturi et al. [40] introduced a kernel function based on Global Alignment Kernel(GAK) - Triangular Global Alignment Kernel (TGAK). However, DTW this distance cannot be easily transformed into a positive definite kernel, because it relies on an optimal calculation, rather than on the construction of feature maps, it cannot be directly utilized as a kernel function. Cuturi proved GAK to be a definite positive nucleus, so we chose GAK to conduct experiments.

Suppose that the $\boldsymbol{x}$ and $\boldsymbol{y}$ are finite time series of length *m* and *n*, respectively, that is, $\boldsymbol{x} = (x_1, x_2, \dots, x_m)$, $\boldsymbol{y} = (y_1, y_2, \dots, y_n)$, and $\boldsymbol{\pi}$ is the length of $p \le m + n - 1$ measures the normalized path distance of two-time series $\boldsymbol{x}$ and $\boldsymbol{y}$, ie. $\boldsymbol{\pi} = (\pi_1, \pi_2)$, so that:

$$\begin{cases} 1 = \pi_1(1) \le \cdots \le \pi_1(p) = m \\ 1 = \pi_2(1) \le \cdots \le \pi_2(p) = n \end{cases} \tag{12}$$

make $\pi_1(i)$ and $\pi_2(j)$ monotonically increasing, $i$, $j \in [0, p-1]$. And,

$$\begin{cases} \pi_1(i+1) \le \pi_1(i) + 1 \\ \pi_2(j+1) \le \pi_2(j) + 1 \\ \pi_1(i+1) - \pi_1(i) + \pi_2(i+1) - \pi_2(i) \ge 1 \end{cases} \quad (13)$$

So that $\pi_1(i)$ and $\pi_2(j)$ do not repeat at the same time. Let $A(m,n)$ be the set of normalized path distances between $\boldsymbol{x}$ and $\boldsymbol{y}$ for time series of lengths *m* and *n*. So the **DTW** distance can be defined as:

$$DTW(\boldsymbol{x},\boldsymbol{y}) = \min_{\pi \in A(m,n)} \sum_{i=1}^{|\pi|} \varphi\left(x_{\pi_1(i)}, y_{\pi_2(i)}\right) \quad (14)$$

where $\varphi$ is defined as the Euclidean distance $\varphi(x,y) = ||x-y||^2$, DTW distance calculates all alignment distances and selects the smallest distance as the similarity measure for time series $\boldsymbol{x}$ and $\boldsymbol{y}$.

Cuturi considers all alignment distances to build the kernel function:

$$k_{GA}(\boldsymbol{x},\boldsymbol{y}) = \sum_{\pi \in A(m,n)} e^{-\sum_{i=1}^{|\pi|} \varphi\left(x_{\pi_1(i)}, y_{\pi_2(i)}\right)} = \sum_{\pi \in A(m,n)} \prod_{i=1}^{|\pi|} k\left(x_{\pi_1(i)}, y_{\pi_2(i)}\right) \quad (15)$$

where $k = e^{-\varphi}$.

Cuturi et al. extended GAK to obtain triangular global alignment kernel TGAK. TGAK acts as a further constraint by introducing a triangular parameter T within GAK. TGAK is not only positive definite; it is faster to compute than GAK. Formula (16) is the structure of TGAK:

$$k_{TGA}(i,x;j,y) = \frac{\omega(i,j) k_\sigma\left(x_i, y_j\right)}{2 - \omega(i,j) k_\sigma\left(x_i, y_j\right)} \quad (16)$$

where $\omega(i,j) = \max\left[\left(1 - \frac{|i-j|}{T}\right), 0\right]$, T>0, $k_\sigma$ is the radial basis kernel function $k_\sigma(x,y) = e^{-\frac{||x-y||^2}{2\sigma^2}}$. The triangular kernel function $\omega(i,j)$ uses the positions $i$ and $j$ of $x_i$ and $y_j$ to modulate the similarity between two points $(x_i, y_j)$, while the radial basis kernel function is used to quantify the similarity of $x_i$ and $y_j$. The triangular parameter T is used to control the zone nearby the diagonal of the kernel matrix. That is the kernel function value in the area farthest from the diagonal is 0 to reduce the amount of computation.

Therefore, we choose TGAK to replace the RBF kernel function of OCKELM, namely TGAK-OCELM, and Algorithm 2 gives the pseudocode of the TGAK-OCELM model.

### *3.2 FastICA*

The FastICA algorithm is widely used in signal processing, mainly for signal separation and denoising, and can also be used for feature extraction of data.

**Algorithm 2 TGAK-OCELM**

Enter:

Training set: **T**, Test instance: $x_t$, Regularization parameter: *C*, Radial basis function parameter: $\sigma$, Outlier scale: $\theta$, Triangle parameter: T

Output:

Label corresponding to $x_t$

(1) Calculate the kernel matrix $\Omega$ by the TGA kernel function $k_{TGA}(i,x;\ j,y) = \frac{\omega(i,j)k_\sigma(x_i,y_j)}{2-\omega(i,j)k_\sigma(x_i,y_j)}$.
(2) The model output ***Otr*** of the training set is calculated by Formula (8).
(3) The abnormality threshold $\delta$ is calculated by Formula (9).
(4) Calculate the kernel function $\begin{bmatrix} K_{TGA}(x_t,x_1) \\ \vdots \\ K_{TGA}(x_t,x_N) \end{bmatrix}^T$ of the test instance $x_t$ with the training set **T** by employing the TGA kernel function $k_{TGA}(i,x;j,y) = \frac{\omega(i,j)k_\sigma(x_i,y_j)}{2-\omega(i,j)k_\sigma(x_i,y_j)}$.
(5) The model output ***Ot*** of the instance $x_t$ is calculated by Equation (10).
(6) The category of the instance $x_t$ is determined by Equation (11).

Suppose a $d$ dimensional random variable $\boldsymbol{x} \in R^{d\times 1}$ is preprocessed and passed through the mixing matrix $\boldsymbol{W} \in R^{d\times d}$ to obtain a new vector $\boldsymbol{S} \in R^{d\times 1}$. The FastICA algorithm maximizes the non-Gaussian property of the vector $\boldsymbol{w}^T\boldsymbol{x}$ ($\boldsymbol{w}$ represents the row vector in the mixture matrix $\boldsymbol{W}$) through a fixed-point iterative algorithm so that s are independent of each other. In order to maximize non-Gaussianness, the most commonly used function to measure non-Gaussianity is the negative entropy approximation. The objective function of the FastICA algorithm is:

$$J_G(w) = [E\{G(\boldsymbol{w}^T\boldsymbol{x})\} - E\{G(v)\}]^2 \tag{17}$$

$G$ represents an arbitrary non-quadratic function, and $v$ represents a random variable with a standard normal distribution. The optimal $\boldsymbol{w}$ is obtained by maximizing $J_G(w)$. To prevent the objective function tends to infinity, Equation (18) is used as the constraint of Equation (17):

$$E((\boldsymbol{w}^T\boldsymbol{x})^2) = E(\boldsymbol{w}^T\boldsymbol{x}\boldsymbol{x}^T\boldsymbol{w}) = E(\boldsymbol{w}^T\boldsymbol{w}) = ||\boldsymbol{w}||^2 = 1 \tag{18}$$

Therefore, the Lagrangian Function of the objective function is:

$$L(\boldsymbol{w}) = E\{G(\boldsymbol{w}^T\boldsymbol{x})\} - \beta||\boldsymbol{w}|| \tag{19}$$

where $\beta$ represents the Lagrange multiplier.

The optimal $\boldsymbol{w}$ is obtained by carrying out the partial derivative concerning $\boldsymbol{w}$ of $L(\boldsymbol{w})$ and setting the derivative equal to 0:

$$E\{\boldsymbol{x}G'(\boldsymbol{w}^T\boldsymbol{x})\} - \beta\boldsymbol{w} = 0 \tag{20}$$

where $G'$ is the derivative of the function $G$. In order to solve $\boldsymbol{w}$ in Equation (20), the FastICA algorithm uses Newton's method. Therefore, let $F(\boldsymbol{w}) = E\{\boldsymbol{x}G'(\boldsymbol{w}^T\boldsymbol{x})\} - \beta\boldsymbol{w}$, and take the partial derivative of $\boldsymbol{w}$ in $F(\boldsymbol{w})$ to get:

$$\begin{aligned} \frac{\partial(E\{\boldsymbol{x}G'(\boldsymbol{w}^T\boldsymbol{x})\} - \beta\boldsymbol{w})}{\partial\boldsymbol{w}} &= E\{\boldsymbol{x}\boldsymbol{x}^{\boldsymbol{T}}G''(\boldsymbol{w}^T\boldsymbol{x})\} - \beta\boldsymbol{I} \\ &\approx \mathrm{E}\{\boldsymbol{x}\boldsymbol{x}^{\boldsymbol{T}}\}\mathrm{E}\{G''(\boldsymbol{w}^T\boldsymbol{x})\} - \beta\boldsymbol{I} \\ &= \mathrm{E}\{G''(\boldsymbol{w}^T\boldsymbol{x})\} - \beta\boldsymbol{I} \end{aligned} \tag{21}$$

Through the above formulas, the iterative form of $\boldsymbol{w}$ is finally obtained by applying Newton's method:

$$\boldsymbol{w}_{n+1} = \boldsymbol{w}_n - \frac{E\{\boldsymbol{x}G'(\boldsymbol{w}_n{}^T\boldsymbol{x})\} - \beta\boldsymbol{w}_n}{\mathrm{E}\{G''(\boldsymbol{w}_n{}^T\boldsymbol{x})\} - \beta\boldsymbol{I}} \tag{22}$$

Transform the Formula (22) to obtain the final iterative Formula of the FastICA algorithm:

$$\boldsymbol{w}_{n+1} = E\{\boldsymbol{x}G'(\boldsymbol{w}_n{}^T\boldsymbol{x})\} - \mathrm{E}\{G''(\boldsymbol{w}_n{}^T\boldsymbol{x})\}\boldsymbol{w}_n \tag{23}$$

The weight vector w obtained by calculating the Formula (23) is normalized:

$$\boldsymbol{w} \leftarrow \frac{\boldsymbol{w}}{||\boldsymbol{w}||} \tag{24}$$

Algorithm 3 gives the pseudocode of the FastICA algorithm:

---

**Algorithm 3** FastICA

---

Enter:

Dataset: $\boldsymbol{x} \in R^{n\times m}$, total iterations: max $_iter$, iterations: $t$, error threshold: $\varepsilon$

Output:

Reconstruction matrix $\boldsymbol{S} \in R^{n\times m}$

(1) Center $\boldsymbol{x}$, where $\boldsymbol{x} = \boldsymbol{x} - \overline{\boldsymbol{x}}, \overline{\boldsymbol{x}}$ is the mean of $\boldsymbol{x}$.
(2) Albino $\boldsymbol{x}$. That is $\widetilde{\boldsymbol{x}} = \boldsymbol{E}\boldsymbol{D}^{-1/2}\boldsymbol{E}^{\boldsymbol{T}}\boldsymbol{x}$ where $E\{\widetilde{\boldsymbol{x}}\widetilde{\boldsymbol{x}}^T\} = \boldsymbol{I}$. $E\{\boldsymbol{x}\boldsymbol{x}^T\} = \boldsymbol{E}\boldsymbol{D}\boldsymbol{E}^T$, $\boldsymbol{E}$, and $\boldsymbol{D}$ are the eigenvector and diagonal eigenvalue matrices, respectively.
(3) Let $t = 1$, and initialize the mixing matrix $\boldsymbol{W} \in R^{n\times n}$.
(4) Update $\boldsymbol{W}$ by Equation (23) and normalize $\boldsymbol{W}$ using Equation (24).
(5) Let $t = t + 1$. If $t \leq$ max $_iter$ and $||\boldsymbol{W}(t+1) - \boldsymbol{W}(t)|| > \varepsilon$, go back to step 4.
(6) Calculate the reconstruction matrix $\boldsymbol{S} = \boldsymbol{W}\boldsymbol{x}$.

---

### *3.3 UAV anomaly data detection process*

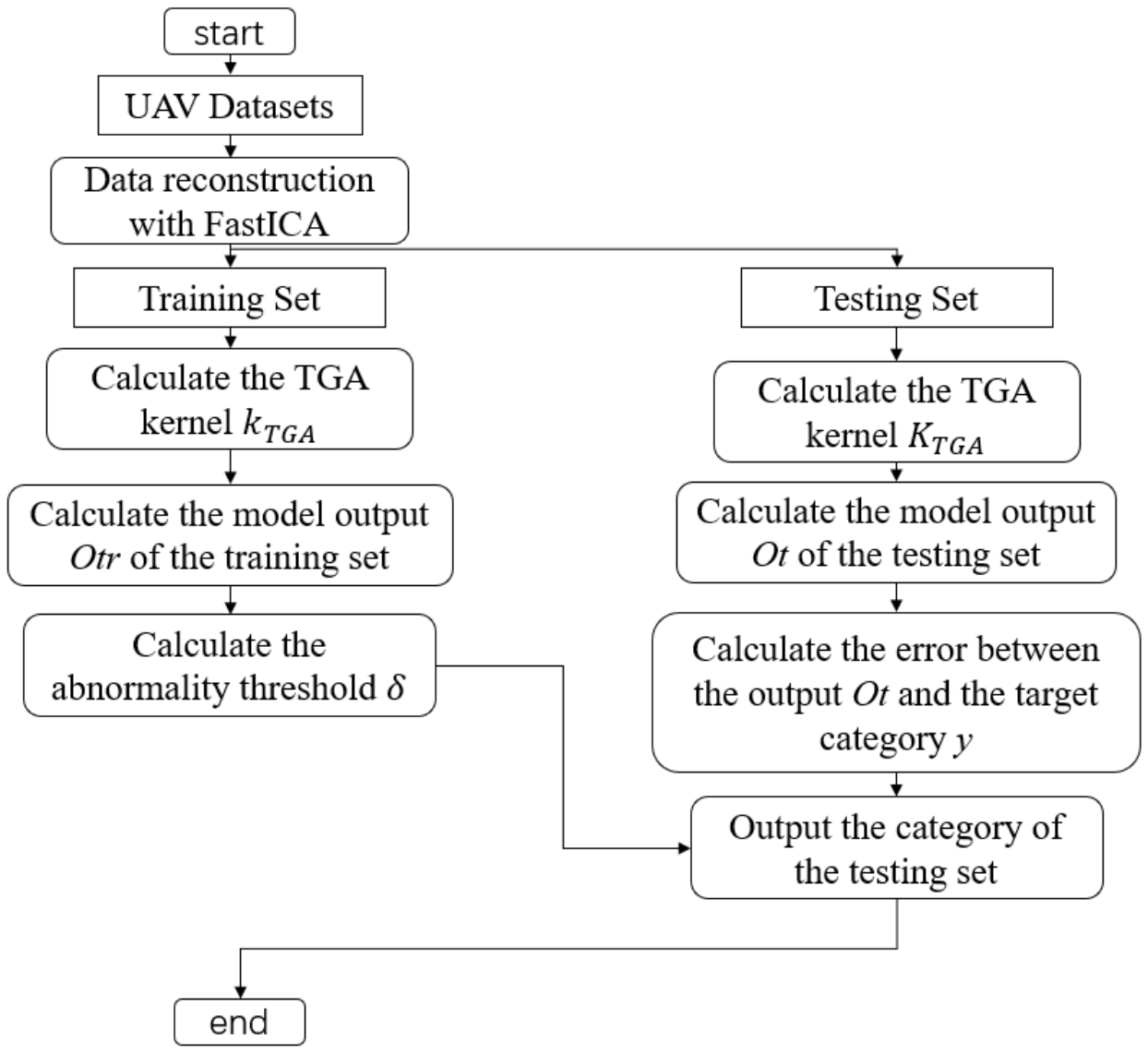

**Figure 1:** UAV anomaly data detection process based on the FastICA-TGAK-OCELM method

This section describes the UAV anomaly data detection process based on the FastICA-TGAK-OCELM method. Algorithm 4 introduces the proposed method. Figure 1 The details of the UAV anomaly data detection process based on the FastICA-TGAK-OCELM method are shown in Figure 1.

---

**Algorithm 4** FastICA-TGAK-OCELM

---

Enter:

Part of the normal flight data of the UAV is used as the training set $\boldsymbol{T} = \{t_1, t_2, \ldots, t_N\}$, and another part of the normal data mixed with abnormal data is used as the test data set X $= \{x_1, x_2, \ldots, x_M\}$, FastICA-TGAK- Relevant hyperparameters of the OCELM model, target label $\boldsymbol{y} = \boldsymbol{1}$.

Output:

Categories of the drone test datasets:

(1) The UAV data is reconstructed by using Algorithm 3.

(2) Calculate the TGA kernel function $\boldsymbol{k}_{TGA}$ by using the training set $\boldsymbol{T}$ and formula (16).

(3) Calculate the model output $\boldsymbol{Otr}$ of the training set $\boldsymbol{T}$ by using equation (8).

(4) The abnormal threshold $\delta$ is calculated by using the formula (9).

(5) Calculate the kernel function $\begin{bmatrix} K_{TGA}(x_t, t_1) \\ \vdots \\ K_{TGA}(x_t, t_N) \end{bmatrix}^T$ of each test instance $x_t$ (t=1,2,...,M) and training set $\boldsymbol{T}$, through the test set $X$ and training set $\boldsymbol{T}$ and formula (16), so we can get the result $\boldsymbol{K}_{TGA} = \begin{bmatrix} K_{TGA}(x_1, t_1) & \ldots & K_{TGA}(x_M, t_1) \\ \vdots & \ddots & \vdots \\ K_{TGA}(x_1, t_N) & \ldots & K_{TGA}(x_M, t_N) \end{bmatrix}^T$ for the test set $X$.

(6) Calculate the model output $\boldsymbol{Ot}$ of the test set X by employing formula (10).

(7) Calculate the error value $\boldsymbol{Dt} = |\boldsymbol{Ot} - \boldsymbol{y}|$ between the output $\boldsymbol{Ot}$ and the target label $\boldsymbol{y}$.

(8) The category of the test set $X$ is determined by deploying formula (11).

## 4 Experiments

### *4.1 Evaluation indicators*

The evaluation metrics utilized to measure the classifier's performance are deployed to evaluate the FastICA-TGAK-OCELM model's classification performance to verify the effectiveness of the FastICA-TGAK-OCELM model. We choose the $F_1$ score as the evaluation index, and the formula of the $F_1$ score is as follows:

$$F_1 = 2 * \frac{\boldsymbol{R} * \boldsymbol{P}}{\boldsymbol{P} + \boldsymbol{R}} \tag{25}$$

The precision rate $\boldsymbol{P} = \frac{TP}{TP+FP}$ represents the number of positive samples in the predicted class of positive samples, and the recall rate $\boldsymbol{R} = \frac{TP}{TP+FN}$ represents the number of positive samples predicted. The category is the proportion of positive samples. $TP$, $FP$, and $FN$ are true positives, false positives, and false negatives, respectively. The predicted class is negative, and the actual class is positive for negative samples.

### *4.2 Hyperparameter settings of the algorithm*

We compare the proposed TGAK-OCELM and FastICA-TGAK-OCELM models with some classical one-class classifiers, namely OCKELM [10], ML-OCKELM [41], OCSVM [42], LOF [43], PCA [44], KNN [45], Isolation Forest [46]. Among them, OCSVM, LOF, PCA, KNN, and Isolation Forest are all from the sklearn [47] library. For the FastICA algorithm, we call sklearn. decomposition.FastICA in the sklearn library to reconstruct the data features. For the hyperparameters involved in FastICA-TGAK-OCELM and TGAK-OCELM: triangular parameter T, regularization parameter $C$, radial basis function parameter $\sigma$, anomaly ratio $\theta$, where the value range of triangular parameter T is $\{2^0, 2^{0.5}, 2^1, \dots, 2^8\}$, the value range of the regularization parameter $C$ is $\{10^{-5}, 10^{-4}, 10^{-3}, \dots, 10^5\}$, the value range of the radial basis function parameter $\sigma$ is $\{2^{-6}, 2^{-5}, 2^{-4}, \dots, 2^6\}$, and the value of the anomaly ratio $\theta$ is set to 0.01. For the hyperparameters involved in the OCKELM algorithm: the regularization parameter $C$ and the radial basis function parameter $\sigma$, the value range of the anomaly ratio $\theta$ is consistent with the value range mentioned earlier. For the hyperparameters involved in the ML-OCKELM algorithm, in addition to the regularization parameter $C$, the radial basis function parameter $\sigma$, and the anomaly ratio $\theta$, which are consistent with the previously mentioned value ranges, the number of layers $q$ of the algorithm should be set to 3 layers.

### *4.3 Experimental evaluation using the UCI dataset*

#### *4.3.1 Data Preprocessing*

We use eight UCI [48] datasets to verify the efficacy of the initiated algorithm before detecting UAV anomaly data. Among them, for multi-category datasets, the dataset is converted into a single-category dataset by selecting one category as the target class (standard data) and other categories as abnormal classes. Since the single-class classifier training set is all the target class samples, the production process of the data set is as follows: the target class and the abnormal class samples are divided into half, and half of the target class is used as the training set and used together with half of the abnormal class samples. Five-fold cross-validation is used to find the best hyperparameters, while the other half of the target class samples and the other half of the abnormal samples are employed as the test set to investigate the model's performance. A elucidation of the dataset is shown in Table 1 [49], which details the dataset size and number of features and selected target classes and enumerates the training and testing set sizes. Before feeding the dataset into the model, we demand to normalize the dataset. We determine zero-mean normalization (z-score normalization) so that the mean of the data is 0, the standard deviation is 1, and the z-score is the conversion formula is:

$$\boldsymbol{x}^* = \frac{\boldsymbol{x} - \overline{\boldsymbol{x}}}{\tau} \tag{26}$$

where $\overline{\boldsymbol{x}}$ represents the mean of the original data $\boldsymbol{x}$, and $\tau$ represents the standard deviation of the original data $\boldsymbol{x}$.

**Table 1:** UCI dataset specification[49]

| Datasets | Total samples | Target | Outlier | Features | Train set | Test set | Target class |
|---|---|---|---|---|---|---|---|
| GlassBuilding | 214 | 76 | 138 | 9 | 38 | 107 | Non float |
| Ionosphere | 351 | 126 | 225 | 34 | 63 | 175 | Bad |
| Iris | 150 | 50 | 100 | 4 | 25 | 75 | Setosa |
| SPECTHeart | 349 | 254 | 95 | 44 | 127 | 174 | Abnormal |
| Biomed | 194 | 67 | 127 | 5 | 34 | 96 | Diseased |
| BreastCancer | 699 | 458 | 241 | 9 | 229 | 349 | Benign |
| Colposcopy | 97 | 82 | 15 | 62 | 41 | 48 | Good |
| Cryotherapy | 90 | 48 | 42 | 6 | 24 | 45 | 1 |

#### *4.3.2 Experimental Results and Discussion*

The experimental results on eight UCI datasets are shown in Figure 2 and Table 2. Figure 2 shows the F1 score of each algorithm on each dataset, from which it can be seen that the two proposed algorithms and other algorithms have obvious differences in the detection effects of GlassBuilding, Ionosphere, SPECTHeart, and Biomed. The effect is slightly improved on the BreastCancer dataset. The detection effect of TGAK-OCELM on the Iris dataset is lower than that of the ML-OCKELM and Isolation Forest algorithms, but the detection effect of the FastICA-TGAK-OCELM algorithm is significantly higher than that of other algorithms. On the Cryotherapy and BreastCancer datasets, the detection effect of the TGAK-OCELM algorithm was slightly higher than that of other algorithms, and the detection effect of the FastICA-TGAK-OCELM algorithm was significantly higher than that of other algorithms. From Table 2, we can more clearly understand the slight gaps in F1 scores of these algorithms. It can be seen from Table 2 that the F1 score of TGAK-OCELM is higher than that of other algorithms on most datasets, and after the introduction of the FastICA method, the results are obtained on six datasets. The F1 score has been enhanced. The F1 score of the FastICA-TGAK-OCELM and TGAK-OCELM methods on the SPECTHeart dataset is not much different. On the Colposcopy dataset, the F1 score of the FastICA-TGAK-OCELM method is 0. The reason is that the Colposcopy dataset is divided into the dataset size after the training set and test set is lower than the number of data features. When utilizing the FastICA algorithm to reduce the dimensionality of the Colposcopy dataset to 10, 20, and 30 dimensions, the F1 scores obtained are 1.0, 0.975, and 0.975, respectively. Three more intuitively show the F1 scores obtained by the FastICA-TGAK-OCELM method under different dimensions of the Colposcopy dataset.

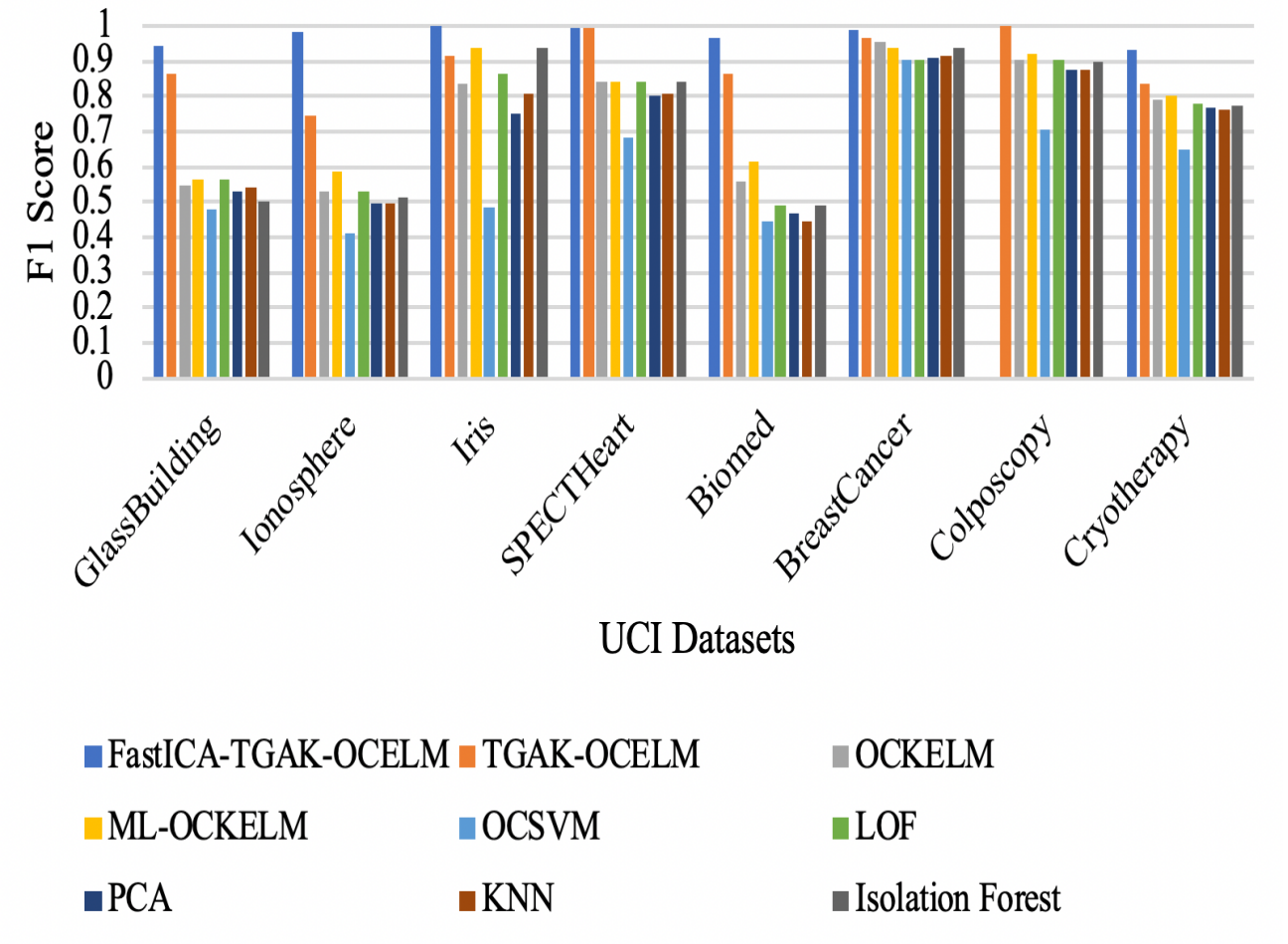

**Figure 2:** Comparison of F1 scores of algorithms on different data

**Table 2:** F1 score comparisons on datasets

| | Glass-Building | Ionosphere | Iris | SPECTHeart | Biomed | BreastCancer | Colposcopy | Cryotherapy |
|---|---|---|---|---|---|---|---|---|
| FastICA-TGAK-OCELM | **0.94444** | **0.98387** | **1** | 0.992063 | **0.96875** | **0.99119** | 0 | **0.93333** |
| TGAK-OCELM | **0.865672** | **0.745562** | **0.913043** | **0.99605** | **0.866667** | **0.963801** | **1** | **0.833333** |
| OCKELM | 0.549451 | 0.529412 | 0.837209 | 0.843854 | 0.556701 | 0.952809 | 0.904762 | 0.792453 |
| ML-OCKELM | 0.563107 | 0.589372 | 0.93617 | 0.843854 | 0.615385 | 0.935185 | 0.921348 | 0.8 |
| OCSVM | 0.481928 | 0.409756 | 0.484848 | 0.682927 | 0.447368 | 0.901679 | 0.705882 | 0.648649 |
| LOF | 0.564516 | 0.529412 | 0.863636 | 0.84 | 0.488189 | 0.904665 | 0.904762 | 0.777778 |
| PCA | 0.5321<br>1 | 0.49785<br>4 | 0.75 | 0.8 | 0.4655<br>17 | 0.912114 | 0.878049 | 0.769231 |
| KNN | 0.5420<br>56 | 0.49785<br>4 | 0.8095<br>24 | 0.808219 | 0.4462<br>81 | 0.914692 | 0.875 | 0.763636 |
| Isolation Forest | 0.5034<br>97 | 0.51282<br>1 | 0.9361<br>7 | 0.843854 | 0.4881<br>89 | 0.935818 | 0.896552 | 0.77193 |

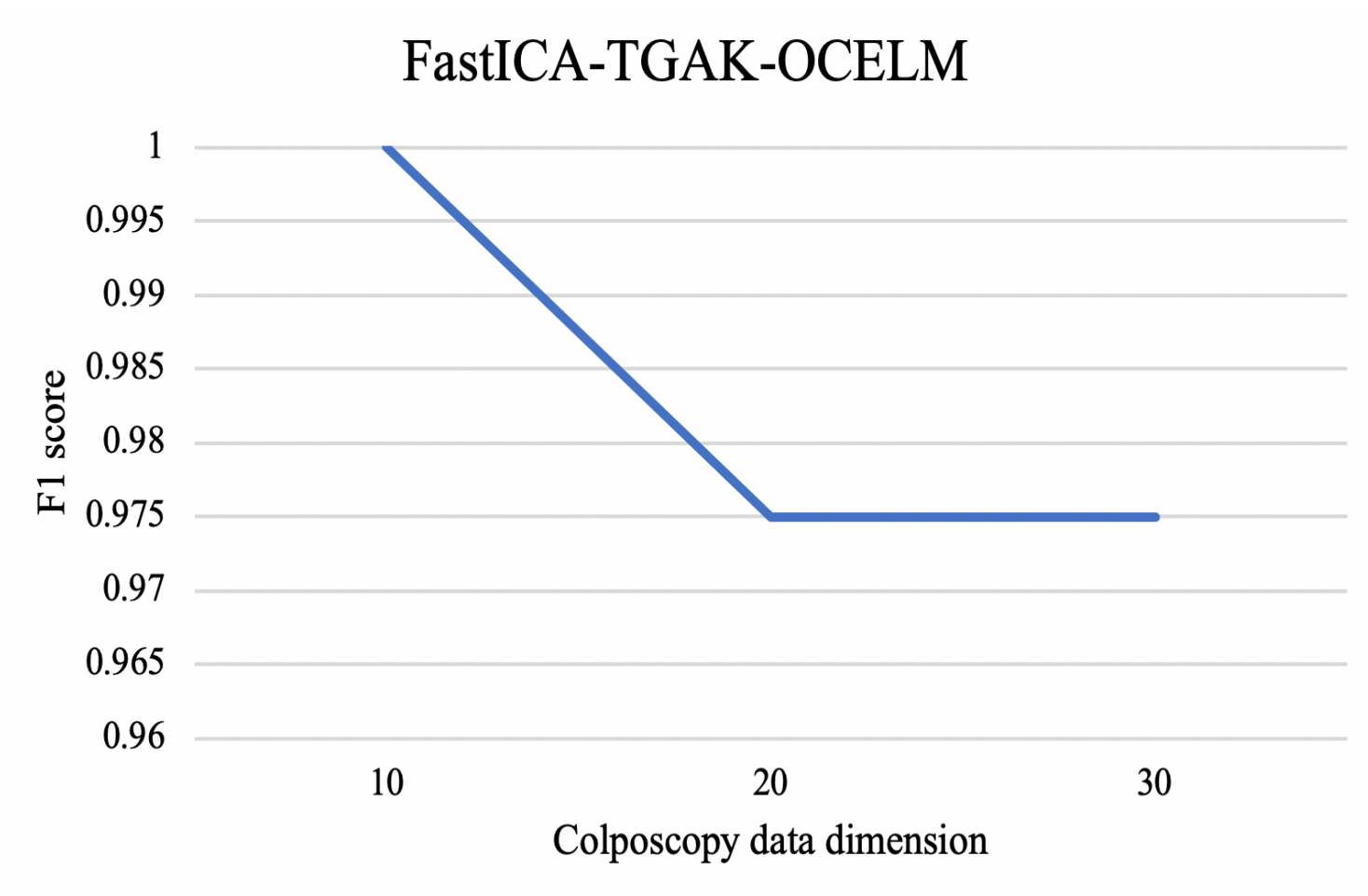

**Figure 3:** F1 scores of FastICA-TGAK-OCELM under different dimensions of Colposcopy dataset

### *4.4 Detecting abnormal drone data*

#### *4.4.1 UAV dataset description and preprocessing*

We handpicked the Air Laboratory Failure and Anomaly (ALFA) [50] dataset to validate the propounded mechanism. This dataset is a flight log generated by a fixed-wing drone performing a circular flight over an airport in Pittsburgh, USA. Fixed-wing UAV (shown in Figure 4) is one of the many types of UAVs. Its structure mainly consists of the fuselage, ailerons, tail, and engine, while the tail is composed of a vertical and horizontal tail, respectively. Controlled by a rudder and elevator. The aileron is used to change the drone's attitude, the rudder changes the orientation of the nose, the elevator makes the fuselage look up and down to rise and fall, and the engine is used to power the drone. In the flight log, there are four failure scenarios of the rudder, ailerons, elevators and engines, and one scenario in which the drone is in complete normal flight. In those four scenarios, the creator of the dataset first let the drone fly normally for a period of time, then controlled the corresponding drone component to make it malfunction, and kept the drone in the flying state, so that no one was generated in the log. The data of the normal flight of the aircraft and the flight data after the failure of the drone.

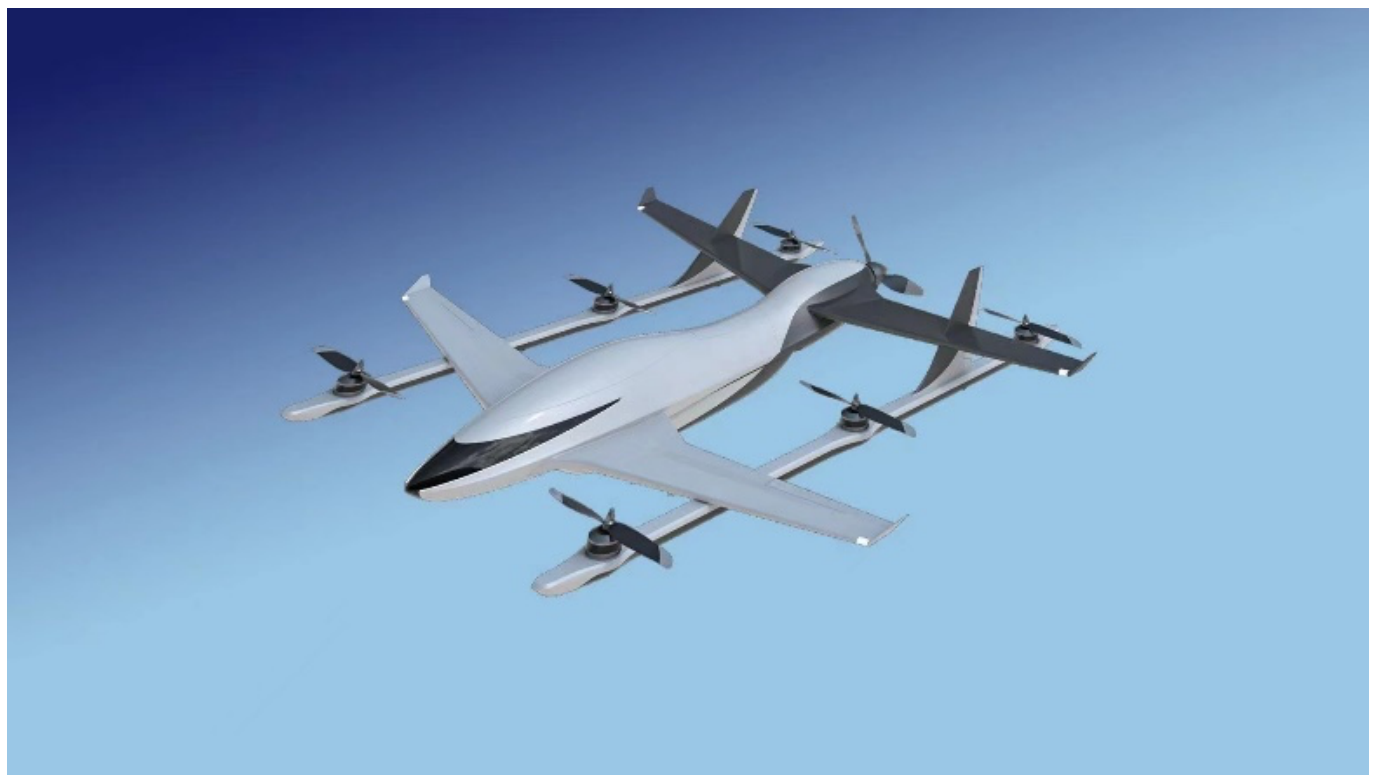

**Figure 4:** Vertical take-off and landing fixed-wing UAV

Since the ALFA dataset is the original flight log, it contains many types of data, such as data associated to the UAV position, data on the UAV system status, data collected by the UAV from the outside world, etc., and the UAV is in the same Different data features are recorded in the period. Each data feature contains different data points. Therefore, the algorithm cannot directly perform abnormal data detection on the ALFA dataset. Therefore, the ALFA dataset needs to be preprocessed.

The ALFA dataset provides timestamps when the drone fails, so we label the data before the drone fails as safe data and the data after it as abnormal. According to the scheme provided by Park [30], the features shown in Table 3 [30] are extracted from the ALFA dataset as the model's input, considering the generality of UAV hardware and excluding features with constant or null values. As shown in Table 3, Velocity (x, y, z) is used to represent the forward direction of the UAV; Angular Velocity (x, y, z), Linear Acceleration (x, y, z), Magnetic Field (x, y) , z) and Fluid Pressure are extracted from the internal measurement unit (IMU) to represent the state of the IMU. The five characteristics of Temperature, Altitude Error, Airspeed Error, Tracking Error (x), and WP Distance are used to represent the system status. The features shown in Table 3 represent the general characteristics of UAVs.

**Table 3:** Feature Names

| Feature Name | Description |
|---|---|
| Velocity (x, y, z) | Measured velocity of axis x, y, and z, respectively |
| Angular Velocity (x, y, z) | Angular velocity at axis x, y, and z, respectively |
| Linear Acceleration (x, y, z) | Linear acceleration at axis x, y, and z, respectively |
| Magnetic Field (x, y, z) | Value of the magnetic field at axis x, y, and z, respectively |
| Fluid Pressure | Value of the pressure using fluid pressure sensors |
| Temperature | The temperature of the battery |
| Altitude Error | Error value of the current altitude |
| Airspeed Error | Error value of current airspeed |
| Tracking Error (x) | Tracking error at x axis |
| WP Distance | Distance between the ideal location and the current location |

The drones recorded different data types in the same period, and the number of data points for different features differed. Figure 5 [30] depicts the number of data points for different features in the same period. In response to this problem, we first set 0.25 seconds to split the timestamp (the timestamp is expressed as the entire flight period of the drone). Since the data is standard data before a certain time point and abnormal data after that, the timestamp is divided into two parts and then split separately. For example, in a 2-second timestamp, the drone starts to fail at 1.2 seconds, then the timestamp is split into {0, 0.25, 0.5, 0.75, 1, 1.2, 1.45, 1.7, 1.95, 2}. Then randomly select a data point in each period to represent the feature point of this period. If there is no data point in this period, copy the data point in the previous period.

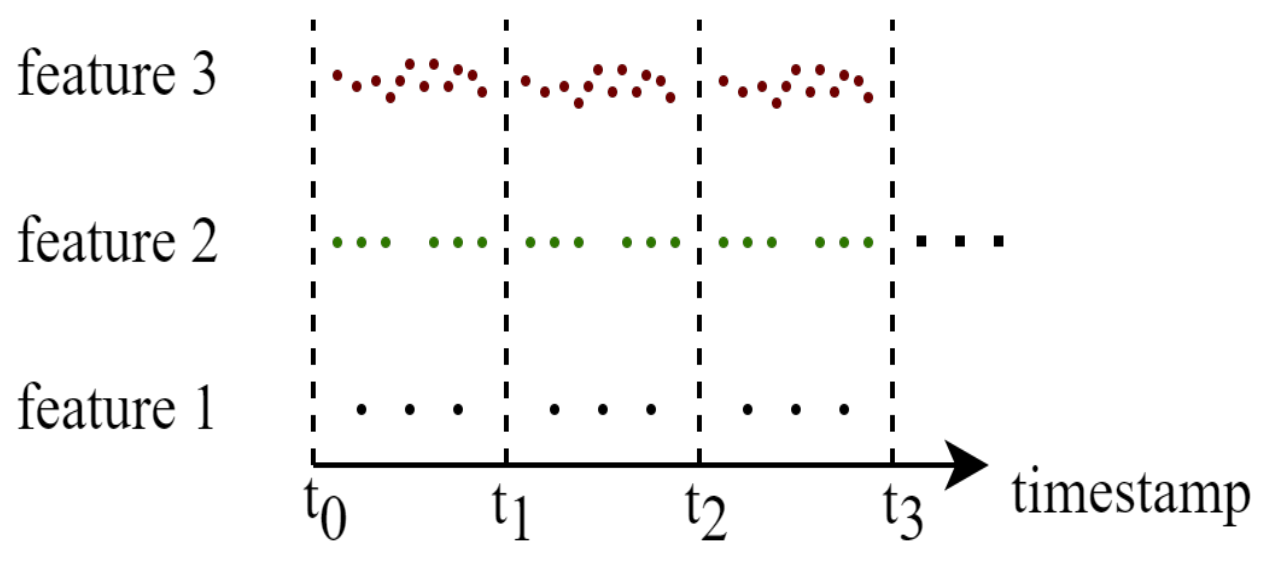

**Figure 5:** Number of data points for different features in the same period

After preprocessing the ALFA data set, it is divided into four data sets according to the fault type, the size of the data set is shown in Table 4, and then the data of these four data sets are normalized by formula (26).

**Table 4:** Specifications of datasets made by fault type

| Fault Type | Total samples | Target | Outlier | Features | Train set | Test set |
|---|---|---|---|---|---|---|
| engine_failure | 1878 | 1630 | 248 | 18 | 815 | 938 |
| ailerons_failure | 3820 | 2528 | 1292 | 18 | 1264 | 1909 |
| elevator_failure | 820 | 726 | 94 | 18 | 363 | 410 |
| rudder_failure | 840 | 672 | 168 | 18 | 336 | 420 |

*4.4.2 Experimental Results and Discussion*

The experimental results of each algorithm in the dataset of four fault types are shown in Figure 6 and Table 5. Figure 6 shows that the detection performance of several algorithms on the rudder_failure dataset is not much different, while in several other faults, the performance of the TGAK-OCELM algorithm is more robust than other algorithms on different types of data sets. It can be seen from Table 5 that the accuracy of the TGAK-OCELM algorithm is more than 2% higher than that of other classic anomaly detection algorithms. After the construction, the FastICA-TGAK-OCELM algorithm was used to improve the detection performance further. Compared with the TGAK-OCELM algorithm, the accuracy was improved by about 0.6%~3%, and the abnormal situation of the UAV data was effectively detected. Improved drone safety.

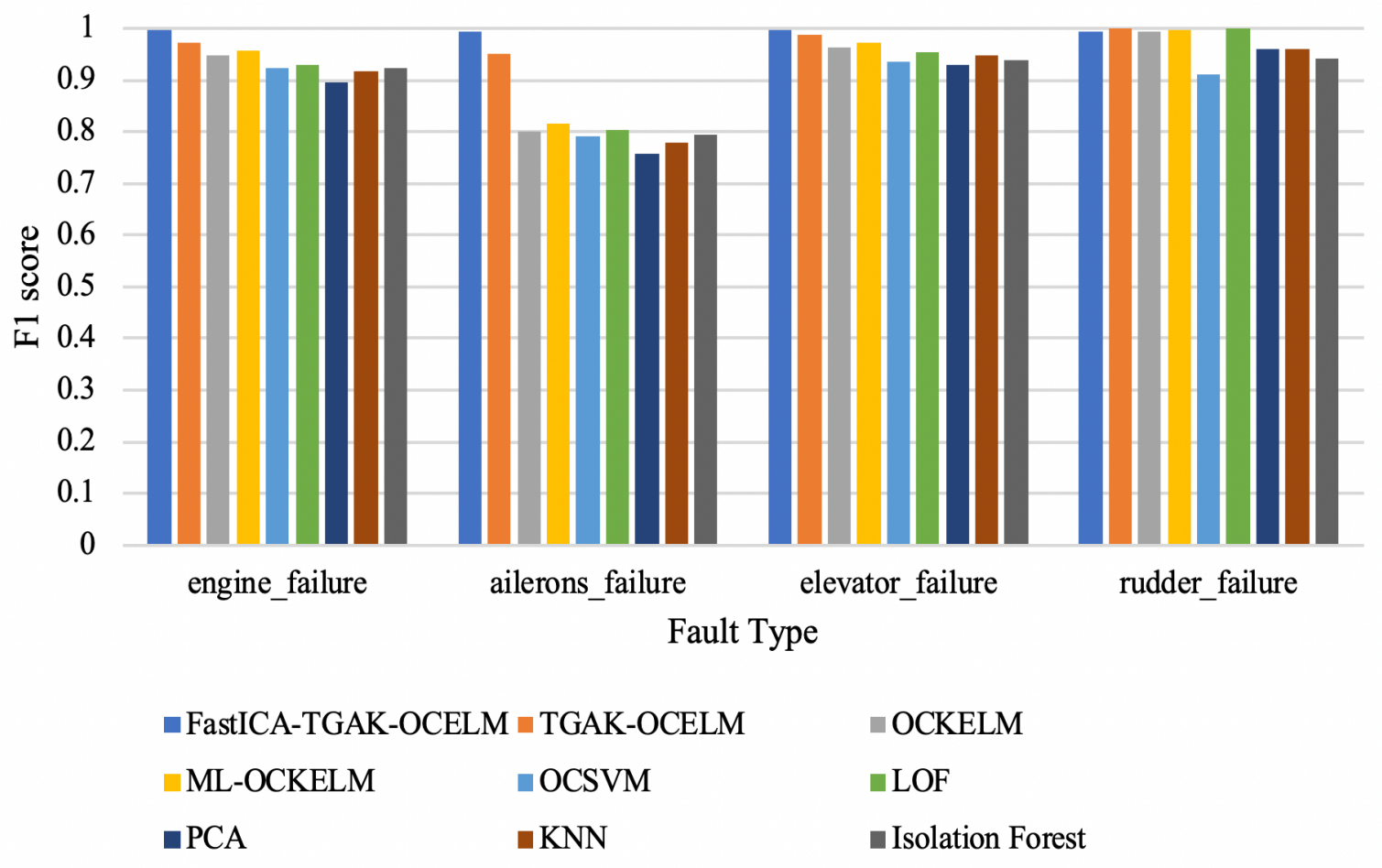

**Figure 6:** Comparison of F1 scores of algorithms on different fault-type datasets

**Table 5:** F1 score comparisons on ALFA datasets

| | engine_failure | ailerons_failure | elevator_failure | rudder_failure |
|---|---|---|---|---|
| FastICA-TGAK-OCELM | **0.99568** | **0.99523** | **0.99585** | 0.994012 |
| TGAK-OCELM | 0.970923 | 0.951982 | 0.988858 | **0.99851** |
| OCKELM | 0.949359 | 0.800901 | 0.963563 | 0.992526 |
| ML-OCKELM | 0.955542 | 0.816063 | 0.972973 | 0.995529 |
| OCSVM | 0.921839 | 0.792536 | 0.936281 | 0.912088 |
| LOF | 0.930854 | 0.803215 | 0.954787 | **0.99851** |
| PCA | 0.894479 | 0.756902 | 0.927978 | 0.96136 |
| KNN | 0.918367 | 0.779273 | 0.947518 | 0.96136 |
| Isolation Forest | 0.923077 | 0.792918 | 0.939948 | 0.941679 |

## 5 Conclusion

To detect the abnormal data of UAVs, we propose a method to detect the data of UAVs. We propose the FastICA-TGAK-OCELM method. Firstly, the UAV data is reconstructed by the FastICA method, and then the TGA core and OCELM are combined to obtain the TGAK-OCELM model. Then, the TGAK-OCELM model is instructed via employing the UAV's standard data, and the trained model is utilized to test the UAV data set. Experimental results show that the proposed algorithm can effectively detect abnormal data. In order to prove the effectiveness of our algorithm, we use the UCI data set to verify the algorithm. The experimental results indicate that our algorithm performs better than other machine learning algorithms.

**Funding Statement:** This article is supported by The Natural Science Foundation of the Jiangsu Higher Education Institutions of China (Grants No.19JKB520031).

**Conflicts of Interest:** The authors declare that they have no conflicts of interest to report regarding the present study.